\documentclass[conference]{IEEEtran}
\IEEEoverridecommandlockouts
\usepackage{cite}
\usepackage{amsmath,amssymb,amsfonts}
\usepackage{algorithmic}
\usepackage{graphicx}
\usepackage{textcomp}
\usepackage{xcolor}
\usepackage{multirow}
\def\BibTeX{{\rm B\kern-.05em{\sc i\kern-.025em b}\kern-.08em
    T\kern-.1667em\lower.7ex\hbox{E}\kern-.125emX}}
\begin{document}

\title{Multi-Source Deep Domain Adaptation for Quality Control in Retail Food Packaging}

\author{\IEEEauthorblockN{ Mamatha Thota \& Stefanos Kollias}
\IEEEauthorblockA{\textit{School of Computer Science} \\
\textit{University of Lincoln}\\
Lincoln, UK \\
\{mthota, skollias\}@lincoln.ac.uk}
\and
\IEEEauthorblockN{ Mark Swainson}
\IEEEauthorblockA{\textit{National Centre for Food Manufacturing} \\
\textit{University of Lincoln}\\
Holbeach Technology Park, UK \\
mswainson@lincoln.ac.uk}
\and
\IEEEauthorblockN{Georgios Leontidis}
\IEEEauthorblockA{\textit{Department of Computing Science} \\
\textit{University of Aberdeen}\\
Aberdeen, UK \\
geleonti@hotmail.com}

}

\maketitle

\begin{abstract}
Retail food packaging contains information which informs choice and can be vital to consumer health, including product name, ingredients list, nutritional information, allergens, preparation guidelines, pack weight, storage and shelf life information (use-by / best before dates). The presence and accuracy of such information is critical to ensure a detailed understanding of the product and to reduce the potential for health risks. Consequently, erroneous or illegible labeling has the potential to be highly detrimental to consumers and many other stakeholders in the supply chain. In this paper, a multi-source deep learning-based domain adaptation system is proposed and tested to identify and verify the presence and legibility of use-by date information from food packaging photos taken as part of the validation process as the products pass along the food production line. This was achieved by improving the generalization of the techniques via making use of multi-source datasets in order to extract domain-invariant representations for all domains and aligning distribution of all pairs of source and target domains in a common feature space, along with the class boundaries. The proposed system performed very well in the conducted experiments, for automating the verification process and reducing labeling errors that could otherwise threaten public health and contravene legal requirements for food packaging information and accuracy. Comprehensive experiments on our food packaging datasets demonstrate that the proposed multi-source deep domain adaptation method significantly improves the classification accuracy and therefore has great potential for application and beneficial impact in food manufacturing control systems.

\end{abstract}


\section{Introduction}
Europe's food and drink sector employs 4.57 million people and has a turnover of €1.1 trillion, making it the largest manufacturing industry in the EU (Source:  Data \& Trends. EU Food \& Drink Industry 2018. FoodDrink Europe 2018). The priority of all nations is to protect and feed their citizens. To assure public health, food safety is a legal requirement across the food supply chain. As part of this control approach all pre-packaged food products are required to display mandatory information on the food pack label. This information serves to ensure that the consumer can make a clear and informed choice as to the nature of their food purchases and are warned of any particular issues which could affect their health, e.g. Product shelf life information, the presence of allergens and also other warnings such as the potential for presence of physical or microbiological hazards associated with the particular food type. Labeling mistakes or legibility issues can therefore create major food safety problems including: Food poisoning, due to the consumption of a product that has exceeded its actual use-by date (due to the date on the packaging either being incorrect or illegible); The triggering of an allergic reaction in a consumer who is susceptible to a particular food allergen, but due to a label fault was not aware that this allergen was present in their chosen food. 

Food product traceability is a legal requirement in the EU,  typically achieved by the presence of production date, time and process line code information on the food packaging. The presence of this information ensures that in the event of an emergency it is possible to identify and remove the affected food products in the supply chain. As a result, any fault in the accuracy or legibility of the pack traceabilility information will result in the supply chain stakeholders having to recall far more product than necessary due to the fact that the specific batch of product actually affected cannot be individually identified \cite{pearson2019distributed}. Such circumstances  result in inefficient food recall processes  and an unnecessarily high level of food waste. All very financially expensive to the food manufacturers both in monetary terms and reputation.  The environmental impact of such events also cannot be overlooked in terms of the impact of food waste on sector carbon footprints and ultimately in exacerbating climate change.

A common approach to overcome these labeling risks in the food supply chain is to manually read and verify the use by dates on the pre-packaged products during the manufacturing process.  These Quality Assurance (QA) checks are typically conducted by an operator, but as such practices are very laborious and repetitive in nature, it places the operator in an error-prone working environment. Another common approach is to use Optical Character Verification (OCV) approaches, where a supervisory system has the correct date code format which gets transferred both to the printer and the vision system. The vision system then verifies the date, heavily relying on the consistency in the date format, packaging and camera view angle, which is often hard to achieve in the food and drink manufacturing environment.
OCV systems also require accurately labeled data to be utilized for training; but labeling a large number of target samples is overly laborious and a very cost-ineffective process, hence the need for a more robust solution. Previous studies to consider deep learning (DL) techniques for OCV have primarily focused on one domain and/or using transfer learning to enhance the performance and generalization of the developed techniques \cite{ribeiro2018end,suh2019robust, ribeiro2018adaptable,katyal2019automated, ribeiro2019deep}.

Over the last few years, DL  \cite{lecun2015deep} has been successfully applied to numerous applications and domains due to the availability of large amounts of labeled data, such as computer vision and image processing \cite{esteva2019guide,de2019capsule,carballo2019new, khan2019novel, kollias2017adaptation}, signal processing \cite{caliva2018deep,zhao2019deep, ribeiro2018towards, ieracitano2019time}, time series analysis and forecasting \cite{langkvist2014review,ONOUFRIOU2019103133,fawaz2019deep,alhnaity2019using, kashiparekh2019convtimenet}, and autonomous driving \cite{wang2019deep,ghosh2019segfast}. As most of the applications of DL techniques, such as the aforementioned ones, refer to supervised learning, labeling large number of datasets consume a lot of time and is very cost ineffective. In addition, when deploying a trained model to real-life applications the assumption is that both the source (training set) and the target one (application-specific) are drawn from the same distributions. When this assumption is violated, the DL model trained on the source domain will not generalize well on the target domain  due to the distribution differences between the source and the target domains known as domain shift.

Learning a discriminative model in the presence of domain shift between source and target datasets is known as \textit{Domain Adaptation}. Typically, domain adaptation methods are used to model labeled data from one single source domain to another called target domain. In real life applications, data from multiple sources and domains do exist which could be leveraged to develop a more robust and generalizable model. The information extracted from these multiple sources will be able to better fit the target distribution data that might not match any of the available source data -- so it is more valuable in the performance improvement and is therefore receiving considerable attention in the real world applications like those described in this paper.

One simple approach for predicting the labels of the target domain data is to combine the training samples from all source domains and build a single model based on the pooled training samples.  Due to the data expansion, the methods might improve the performance, however, this simple approach will not work well in our application as there are significant conditional probability differences across domains in the food image data. Another approach is to extract multiple domain-invariant representations for each source and target domain pairs and map each of them into specific feature spaces in order to match their distributions, by training multiple models. However, this would take a lot of time as it involves training multiple models, therefore it is necessary to find a better way to make full use of multiple source domains. So here we propose an approach to overcome the above mentioned issues. 
To the best of our knowledge this is the first study to consider multi-source deep learning domain adaptation in the retail food packaging control, showing the great potential of advancing related quality assurance systems in the food supply chain, further supporting automation towards industry 4.0 and with high potential to reduce errors and their related costs to the consumer and food business operators.

The rest of the paper is laid out as follows: Section II presents the related work in single source and multi source domain adaptation. Section III describes the dataset and our proposed approach, focusing on multi-source domain adaptation techniques, Section IV presents the experimental results obtained after applying our model to the food packaging data and Section V concludes the paper.

\begin{figure*}[t]
\centering
\centerline{\includegraphics[width=\textwidth]{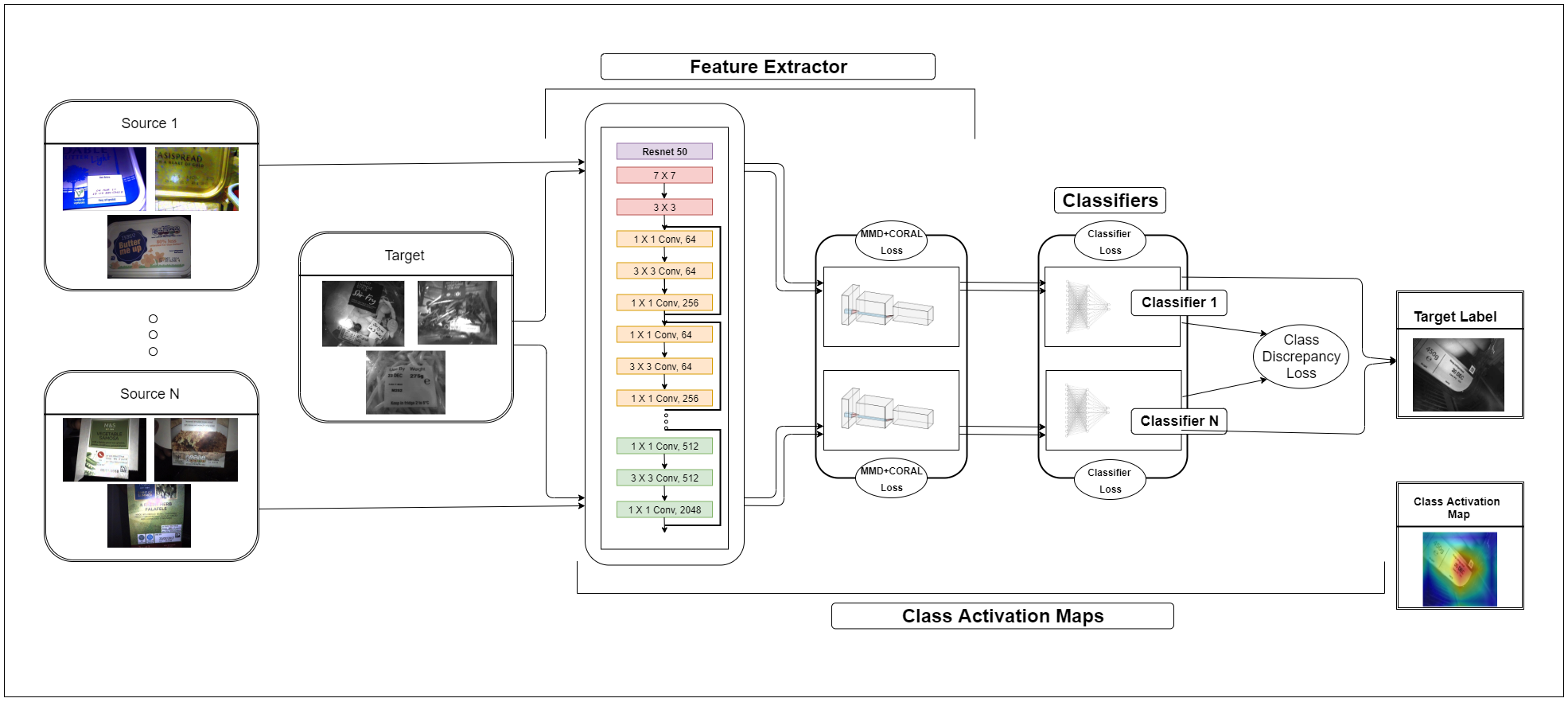}}
\caption{Overview of our proposed Multi-Source Domain Adaptation model.}
\label{fig}
\end{figure*}

\section{Related Work}

In recent years, many single source domain adaptation methods have been proposed. Discrepancy-, Adversarial- and Reconstruction- based approaches are the three primary domain adaptation approaches currently being applied to address the distribution shift \cite{wang2018deep}. Discrepancy-based approaches rely on aligning the distributions in order to minimize the divergence between them. The most commonly used discrepancy-based methods are Maximum Mean Discrepancy (MMD) \cite{long2015learning}, Correlation Alignment (CORAL) \cite{sun2016deep} and Kullback–Leibler divergence \cite{zhuang2015supervised}. Adversarial-based approaches minimize the distance between the source and the target distributions using domain confusion, an adversarial method used in Generative Adversarial Networks. The approach proposed in \cite{ganin2014unsupervised} tries to minimize the feature distributions by integrating a gradient reversal layer, whereas in  \cite{shen2017wasserstein} the aim is to minimize the distance between source and target samples in an adversarial manner using Wasserstein distance. Another class of approaches known as Reconstruction-based approaches create a shared representation between the source and the target domains whilst preserving the individual characteristics of each domain.  Rather than minimizing the divergence, the method in \cite{ghifary2016deep} learns joint representations that classify the labeled source data and at the same time reconstruct the target domain.

In contrast to the single source domain adaptation techniques, multi-source domain adaptation techniques assume the data is available from multiple source domains.  In domain adaptation, multi-source domains are even more critical than single source domain as they need to handle both domain alignment between source and target domains, along with alignments between multiple available sources. The previous single source domain adaptation methods, as the name suggests, only consider a single source and a single target, however in real world applications, there are multiple source domains available to extract knowledge from.

Learning from multiple different sources was originated from early theoretical analysis \cite{ben2010theory},\cite{crammer2007learning}, and has many practical applications. Initially many shallow models were proposed in order to tackle the multi-source domain adaptation problem \cite{jhuo2012robust} \cite{liu2016structure}. The work in \cite{crammer2007learning} established a general bound on the expected loss of the model by minimizing the empirical loss on the nearest k sources. Deep Cocktail Network \cite{xu2018deep} proposed a multi-way adversarial learning to minimize the discrepancy between the target and each of the multiple source domains. The work most related to ours has been the one in \cite{zhu2019aligning}, where only MMD loss was used to minimize the feature discrepancy. Our proposed approach aims at minimizing the feature discrepancy through implementing a new loss function that includes both MMD and CORAL losses in for improved generalization. Additionally, the Class Activation Mapping (CAM) component of our model \cite{zhou2016learning} adds an extra step to the algorithm that provides a visualization of which areas in the image contributed the most to the decision-making process, enabling the trust of the end-users.

\section{Methodology}
\subsection{Problem Statement}

Multi-source domain adaptation with \textit{N} different source distributions is denoted as $\{p_{sj}(x,y)\}^N_{j=1}$, and the labeled source domain image data $\{(X_{sj},Y_{sj})\}^N_{j=1}$ are drawn from these distributions, where $X_{sj} = \{x_{i}^{sj}\}^{cj}_{i=1}$ represents image data sampled with $c$ number of images from source domain \textit{j}  and $Y_{sj} =\{y_{i}^{sj}\}^{cj}_{i=1}$ is the corresponding ground truth labels. We also have the target distribution $p_t(x,y)$, from which the target domain image data $X_t =\{x_{i}^t)\}^{c}_{i=1}$ with a total number of \textit{c} images sampled without any label observation $Y_t$. The end goal is to predict the labels of the unlabeled data in the target domain using the labeled multi-source domain data. 

\subsection{Dataset Description}
\begin{figure}[t]
    \includegraphics[width=\linewidth]{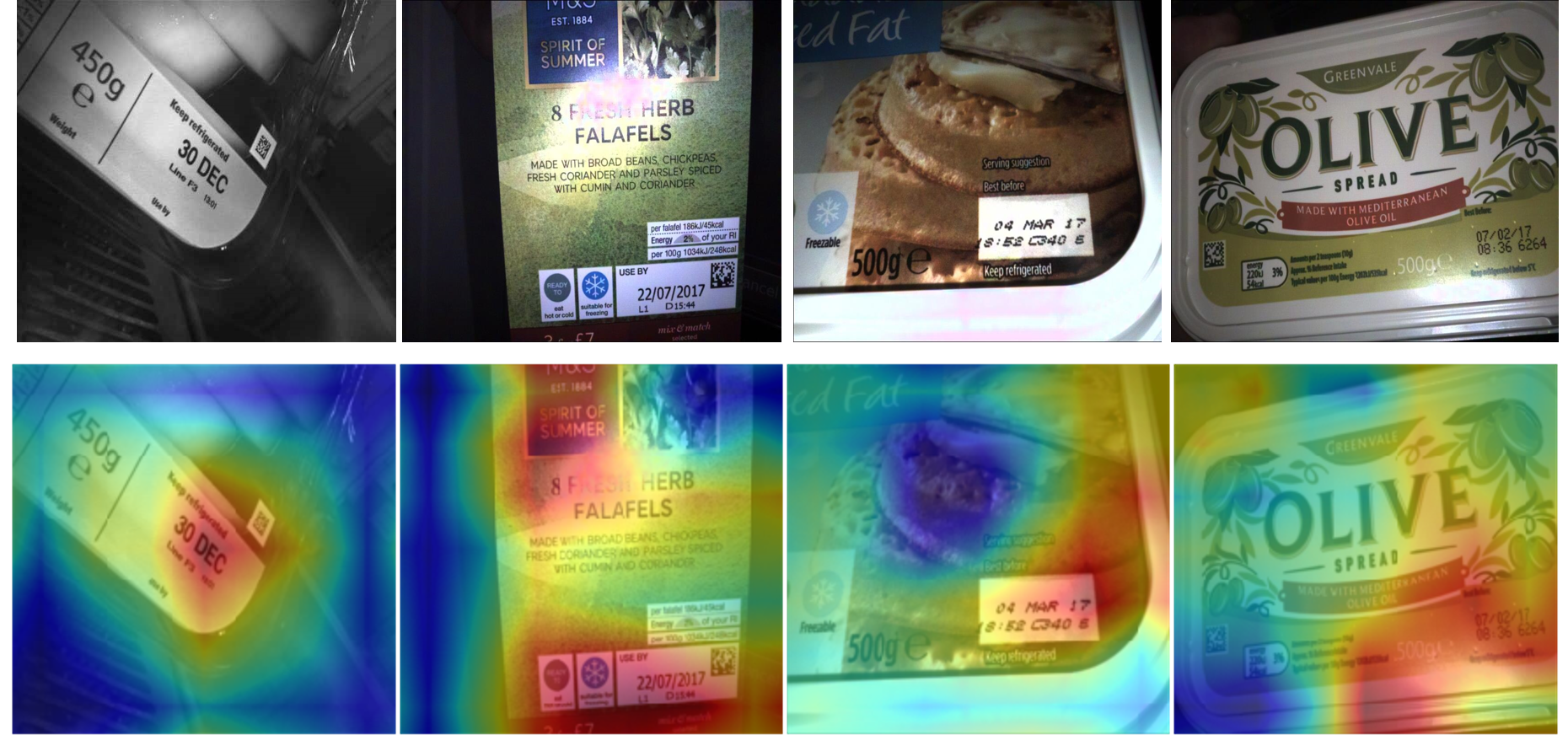}
    \caption{Example images that are considered to be of acceptable quality (OK) for the purposes of our implementations.}
    \label{fig:1}
\end{figure}

\begin{figure}[t]
    \includegraphics[width=\linewidth]{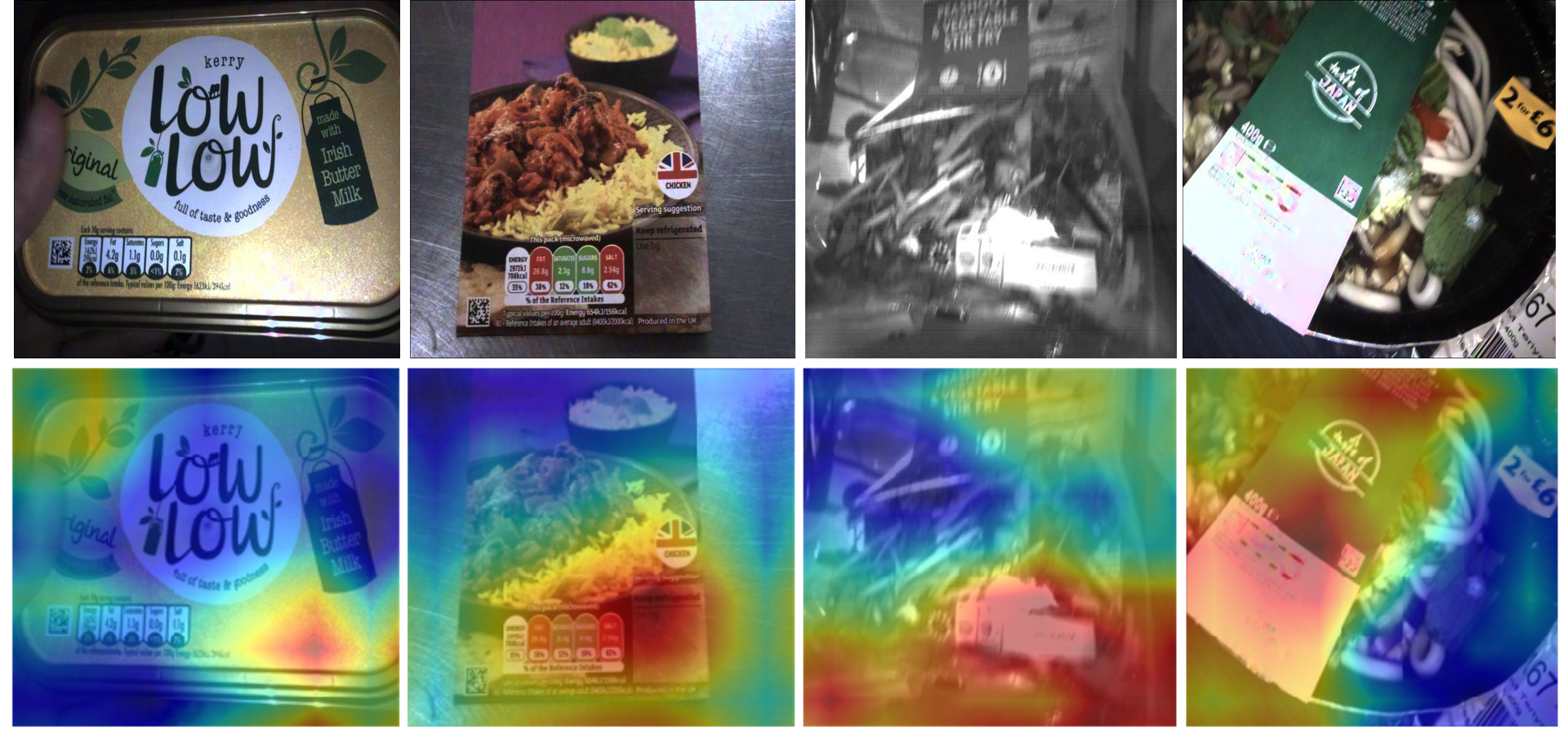}
    \caption{Example images that are considered to be of unacceptable/bad quality (NOT-OK) for the purposes of our implementations.}
    \label{fig:2}
\end{figure}

The Food Packaging Image dataset used in this study consists of more than 30,000 balanced images (OK vs NOT-OK) from six different locations [Abbeydale (Ab), Burton (Bu), Boston (Bo), Listowel (Li), Windmill-Lane (Wi) and Ossett (Os)], and the task is to automatically verify the quality of printed use-by dates. Initially, three people manually annotated this dataset, with two more annotators further sampling and verifying the manual annotations for quality control, hence keeping those 30,000 images that both annotators were in full agreement. The main challenges of these datasets are the unavailability of labeled data and high variability across the datasets, such as heavy distortion, varying background, illumination/blur, date format, angle and orientation of the label, etc. We explore various techniques and propose an approach for adapting knowledge learnt by one dataset to another. The training process was carried out on a 70\% sample with another 10\% used for the validation process. Finally, the remaining 20\% of the images were used for evaluating and testing all the methods across the six distinct locations listed above in order to automatically verify the quality of printed use-by dates, hence detecting images of very low quality. Examples of how the images used in this study look like can be seen in figures \ref{fig:1} and \ref{fig:2}.

\subsection{Overview of our Architecture}
In this section our proposed multi-source DL-based domain adaptation approach is introduced that aims at improving the binary classification of the food packaging dataset across all locations. We use labeled source data from multiple locations and unlabeled target data from a single location. As shown in figure \ref{fig}, our model comprises of a feature extractor and a classification part. The feature extractor part learns useful representations for all domains, whereas its sub-network learns features specific to each source-target domain pairs. The classification part of the model learns domain-specific attributes for each target image and provides $N$ categorization results. The classification part of the model aligns with the domain-specific classifiers’, as the class boundaries are highly likely to be misclassified, because they are learned from different classifiers. The CAM part of the model helps with the visualization of the CNN's interpretation of the model when making predictions.

\subsection{Multi-Source Domain Adaptation}
Domain Adaptation is challenging due to the domain shift between the source and the target datasets. Multi-source domain adaptation is more challenging as it has to deal with domain shift between multiple source datasets. Our model
is composed of a feature extractor and source specific classification parts and aims at minimizing the feature discrepancy, for learning domain-invariant representations, the class boundary discrepancy, for minimizing the mismatch among all classifiers, and finally improving source data classification by reducing the classification loss, leading to improved generalization on the target dataset. 

\textbf{Feature Discrepancy Loss}: We reduce the feature discrepancy by minimizing both MMD and CORAL loss. The proposed deep domain adaptation architecture jointly adapts features using two popular feature adaptation metrics that combine MMD with CORAL in order to align higher order statistics along with the first and second order statistics.

MMD defines the distance between the two distributions with their mean embeddings in the Reproducing Kernel Hilbert Space (RKHS). MMD is a two sample kernel test to determine whether to accept or reject the null hypothesis $p = q$ \cite{gretton2012kernel}, where $p$ and $q$ are source and target domain probability distributions. Let $H$ be the RKHS with a characteristic kernel \textit{k}, the squared distance formulation of MMD is shown in the equation \eqref{1}

\begin{equation} \label{1}
d_k^2(p,q) =|| E_p[\phi (x^s)] -E_q[\phi(x^t)] ||_{H}^2
\end{equation}

The distance between the distributions with their mean embeddings is given in equation \eqref{2}

\begin{equation}\label{2}
Loss_{{MMD}}  = \left\|\frac{1}{N}\sum_{i=1}^N\phi (x^s_i)-\frac{1}{M}\sum_{j=1}^M\phi (x^t_j) \right\|^2
\end{equation}

\begin{equation}\label{3}
\begin{aligned}
= \frac{1}{N^2}\sum_{i=1}^N\sum_{i^\prime=1}^N\phi (x^s_i)^T\phi (x^s_{i^\prime}) - \frac{2}{NM}\sum_{i=1}^N\sum_{j=1}^M\phi (x^s_i)^T\phi (x^t_j)\\ + \frac{1}{M^2}\sum_{j=1}^M\sum_{j^\prime=1}^M\phi (x^t_j)^T\phi (x^t_{j^\prime}) 
\end{aligned}
\end{equation}
where N and M are the total number of items in the source and target respectively.

The kernel trick can be applied as each term in the equation \eqref{3} involves inner products between $\phi$ vectors in order to estimate the squared
distance between the empirical kernel mean embeddings as follows

\begin{equation}\label{4}
\begin{aligned}
Loss_{{MMD}}  = \frac{1}{N^2}\sum_{i=1}^N\sum_{i^\prime=1}^Nk(x^s_i,x^s_{i^\prime})-\frac{2}{NM}\sum_{i=1}^N\sum_{j=1}^Mk(x^s_i,x^t_j) \\+ \frac{1}{M^2}\sum_{j=1}^M\sum_{j\prime=1}^Mk(x^t_j,x^t_{j^\prime}) 
\end{aligned}
\end{equation}

CORAL loss \cite{sun2016deep} is also used to minimize the discrepancy between source and target data by reducing the distance between the source and target feature representations.
We define the CORAL loss as the distance between the second-order statistics (covariances) of the source and target features:

\begin{equation}\label{5}
Loss_{CORAL} = \frac{1}{4d^2}||C_s-C_t||_F^2
\end{equation}

$||.||_F^2$  where denotes the squared matrix Frobenius norm, $C_s$ is the source covariance matrix and $C_t$ is the target covariance matrix.
 
 The covariance matrices of the source and target data are given by:
 
 \begin{equation}\label{6}
     C_s = \frac{1}{N-1}(D_s^TD_s - \frac{1}{N}(1^TD_s)^T(1^TD_s))
\end{equation}

 \begin{equation}\label{7}
    C_t= \frac{1}{M-1}(D_t^TD_t - \frac{1}{M}(1^TD_t)^T(1^TD_t))
\end{equation}
 
1 is a column vector with all elements equal to 1, N and M are the total number of items in the source and target respectively. 
 
The total feature discrepancy loss is therefore given by the equation \eqref{8}

\begin{equation}\label{8}
   Loss_{FD} = Loss_{MMD} + Loss_{CORAL}
\end{equation}

\textbf{Class Discrepancy Loss}: Classifiers are likely to misclassify the target samples near the class boundary as they are trained using different source domains, each having different target prediction. Therefore we aim at minimizing the discrepancy among all classifiers by making their probabilistic outputs similar.  The class discrepancy is calculated by the equation \eqref{9}

\begin{equation}\label{9}
Loss_{CD} = \binom N2^{-1} \sum_{j=1}^{N-1} \sum_{i=j+1}^{N}[|E(X_i) - E(X_j)|]
\end{equation}
where N is total number of classifiers.

\textbf{Classification Loss}: The network reduces the discrepancy among classifiers by minimizing the classification loss. We train the network with labeled source data and calculate the empirical loss through minimizing the cross-entropy loss as follows

\begin{equation}\label{10}
Loss_{CL} = \frac{1}{N}\sum_{i=1}^{N}V(f (x_i^s), y_i^s) 
\end{equation}
where $V( . , . )$ is the cross-entropy loss function and $f (x_i^s)$ is the conditional probability that the CNN assigns to label $y_i^s$.

 Our total loss is made up of classiﬁcation loss (CL), feature discrepancy loss (FD) and class discrepancy loss (CD). By minimizing each of these losses the network can classify the source domain data more accurately and reduce the dataset bias and the discrepancy among classiﬁers. Jointly minimizing these three losses, the network can learn features that generalize and adapt well on the target dataset. The overall objective of our network can be formulated as

 \begin{equation}
  Loss_{Total} = Loss_{CL} + \lambda Loss_{FD} + \gamma Loss_{CD}
\end{equation}
where $\lambda$ and $\gamma$ are penalty parameters.

\subsection{Class Activation Maps}
Deep neural networks are often considered to be black boxes that offer no straightforward way of understanding what a network has learned or which part of an input to the network was responsible for the prediction of the network. When such models make predictions, there is no explanation to justify them. Class activation maps \cite{zhou2016learning} are an efficient way to visualize the importance the model places on various regions in an image while training, offering insights that are crucial for the model’s usability and explainability. We have incorporated CAM to our approach in order to visualize which areas of the food packaging images contribute the most to the decision-making process of the algorithm. 

CAM provides some insight into the process of CNN interpretability and explainability by overlaying a heat map over the original image to demonstrate where the model is paying more attention for its decision-making process. It visually demonstrates how the algorithm comes up with its prediction by highlighting the pixels of the image that trigger the model to associate the image with a particular class.
CAM help us understand which regions of an input image influence the convolutional neural network’s output prediction. Such an information can be used for examining the bias of the algorithm and the lack of generalization capabilities, allowing to take steps to enhance the robustness of the model and potentially increase its accuracy.

\section{Experiments and Results}
\begin{table}[]
\centering
\caption{Comparison of classification accuracy and average accuracy (\%) on food packaging target dataset Ossett ($Os$)}
\label{tab:my-table Os}
\begin{tabular}{|l|l|l|l|}
\hline 
\textbf {Method} & \textbf {Source} & \textbf{Accuracy} & \textbf {Average} \\ \hline
Single-Source & \textbf{Ab} & 84.7 & 84.70 \\ \hline
\multirow{4}{*}{Source Combined with 2 datasets} & Ab,Bo & 85.2 & \multirow{4}{*}{85.30} \\ \cline{2-3}
 & Ab,Bu & 86.1 &  \\ \cline{2-3}
 & Ab,Li & 85.3 &  \\ \cline{2-3}
 & Ab,Wi & 84.6 &  \\ \hline
\multirow{6}{*}{Source Combined with 3 datasets} & Ab,Bo,Wi & 86.2 & \multirow{6}{*}{86.38} \\ \cline{2-3}
 & Ab,Bu,Li & 85.4 &  \\ \cline{2-3}
 & Ab,Bu,Bo & 86.6 &  \\ \cline{2-3}
 & Ab,Bu,Wi & 87.2 &  \\ \cline{2-3}
 & Ab,Li,Bo & 85.6 &  \\ \cline{2-3}
 & Ab,Li,Wi & 87.3 &  \\ \hline
\multirow{4}{*}{Multi-Source with 2 datasets}  & Ab,Bo & 88.7 & \multirow{4}{*}{90.35} \\ \cline{2-3}
 & Ab,Bu & 90.9 &  \\ \cline{2-3}
 & Ab,Li & 89.9 &  \\ \cline{2-3}
 & Ab,Wi & 91.9 &  \\ \hline
\multirow{6}{*}{Multi-Source with 3 datasets} & Ab,Bo,Wi & 93.6 & \multirow{6}{*}{92.61} \\ \cline{2-3}
 & Ab,Bu,Li & 91.5 &  \\ \cline{2-3}
 & Ab,Bu,Bo & 91.8 &  \\ \cline{2-3}
 & Ab,Bu,Wi & 92.7 &  \\ \cline{2-3}
 & Ab,Li,Bo & 93.5 &  \\ \cline{2-3}
 & Ab,Li,Wi & 92.6 &  \\ \hline
\end{tabular}
\end{table}

\begin{table}[]
\centering
\caption{Comparison of classification accuracy and average accuracy (\%) on food packaging target dataset Listowel ($Li$)}
\label{tab:my-table-Li}
\begin{tabular}{|l|l|l|l|}
\hline 
\textbf {Method} & \textbf {Source} & \textbf{Accuracy} & \textbf {Average} \\ \hline
Single-Source & Ab & 86.2 & 86.20 \\ \hline
\multirow{4}{*}{Source Combined with 2 datasets} & Ab,Bo & 84.4 & \multirow{4}{*}{86.92} \\ \cline{2-3}
 & Ab,Bu & 87.4 &  \\ \cline{2-3}
 & Ab,Os & 87.6 &  \\ \cline{2-3}
 & Ab,Wi & 88.3 &  \\ \hline
\multirow{6}{*}{Source Combined with 3 datasets} & Ab,Bo,Wi & 85.2 & \multirow{6}{*}{87.15} \\ \cline{2-3}
 & Ab,Bu,Os & 88.6 &  \\ \cline{2-3}
 & Ab,Bu,Bo & 88.1 &  \\ \cline{2-3}
 & Ab,Bu,Wi & 88.7 &  \\ \cline{2-3}
 & Ab,Os,Bo & 82.1 &  \\ \cline{2-3}
 & Ab,Os,Wi & 90.2 &  \\ \hline
\multirow{4}{*}{Multi-Source with 2 datasets} & Ab,Bo & 89.6 & \multirow{4}{*}{90.87} \\ \cline{2-3}
 & Ab,Bu & 92.1 &  \\ \cline{2-3}
 & Ab,Os & 91.3 &  \\ \cline{2-3}
 & Ab,Wi & 90.5 &  \\ \hline
\multirow{6}{*}{Multi-Source with 3 datasets} & Ab,Bo,Wi & 91.2 & \multirow{6}{*}{92.35} \\ \cline{2-3}
 & Ab,Bu,Os & 92.3 &  \\ \cline{2-3}
 & Ab,Bu,Bo & 92.2 &  \\ \cline{2-3}
 & Ab,Bu,Wi & 92.6 &  \\ \cline{2-3}
 & Ab,Os,Bo & 92.1 &  \\ \cline{2-3}
 & Ab,Os,Wi & 93.7 &  \\ \hline
\end{tabular}
\end{table}

\begin{table}[]
\centering
\caption{Comparison of classification accuracy and average accuracy (\%) on food packaging target dataset Burton ($Bu$)}
\label{tab:my-table Bu}
\begin{tabular}{|l|l|l|l|}
\hline 
\textbf {Method} & \textbf {Source} & \textbf{Accuracy} & \textbf {Average} \\ \hline
Single-Source & \textbf{Ab} & 83.5 & 83.50 \\ \hline
\multirow{4}{*}{Source Combined with 2 datasets} & Ab,Bo & 84.1 & \multirow{4}{*}{84.95} \\ \cline{2-3}
 & Ab,Li & 84.7 &  \\ \cline{2-3}
 & Ab,Os & 88.3 &  \\ \cline{2-3}
 & Ab,Wi & 82.7 &  \\ \hline
\multirow{6}{*}{Source Combined with 3 datasets} & Ab,Bo,Wi & 84.95 & \multirow{6}{*}{86.14} \\ \cline{2-3}
 & Ab,Li,Os & 85.4 &  \\ \cline{2-3}
 & Ab,Li,Bo & 85.1 &  \\ \cline{2-3}
 & Ab,Li,Wi & 84.9 &  \\ \cline{2-3}
 & Ab,Os,Bo & 88.6 &  \\ \cline{2-3}
 & Ab,Os,Wi & 87.9 &  \\ \hline
\multirow{4}{*}{Multi-Source with 2 datasets} & Ab,Bo & 90.6 & \multirow{4}{*}{90.60} \\ \cline{2-3}
 & Ab,Li & 88.7 &  \\ \cline{2-3}
 & Ab,Os & 92.9 &  \\ \cline{2-3}
 & Ab,Wi & 90.2 &  \\ \hline
\multirow{6}{*}{Multi-Source with 3 datasets} & Ab,Bo,Wi & 91.4 & \multirow{6}{*}{92.46} \\ \cline{2-3}
 & Ab,Li,Os & 92.6 &  \\ \cline{2-3}
 & Ab,Li,Bo & 92.9 &  \\ \cline{2-3}
 & Ab,Li,Wi & 91.3 &  \\ \cline{2-3}
 & Ab,Os,Bo & 93.4 &  \\ \cline{2-3}
 & Ab,Os,Wi & 93.2 &  \\ \hline
\end{tabular}
\end{table}

\begin{table}[]
\centering
\caption{Comparison of classification accuracy and average accuracy (\%) on food packaging target dataset Abbeydale ($Ab$)}
\label{tab:my-table Ab}
\begin{tabular}{|l|l|l|l|}
\hline 
\textbf {Method} & \textbf {Source} & \textbf{Accuracy} & \textbf {Average} \\ \hline
Single-Source & \textbf{Bu} & 84.6 & 84.60 \\ \hline
\multirow{4}{*}{Source Combined with 2 datasets} & Bu,Bo & 82.3 & \multirow{4}{*}{85.15} \\ \cline{2-3}
 & Bu,Li & 86.6 &  \\ \cline{2-3}
 & Bu,Os & 83.8 &  \\ \cline{2-3}
 & Bu,Wi & 87.9 &  \\ \hline
\multirow{6}{*}{Source Combined with 3 datasets} & Bu,Bo,Wi & 83.7 & \multirow{6}{*}{86.75} \\ \cline{2-3}
 & Bu,Li,Os & 87.5 &  \\ \cline{2-3}
 & Bu,Li,Bo & 88.2 &  \\ \cline{2-3}
 & Bu,Li,Wi & 87.2 &  \\ \cline{2-3}
 & Bu,Os,Bo & 85.6 &  \\ \cline{2-3}
 & Bu,Os,Wi & 88.3 &  \\ \hline
\multirow{4}{*}{Multi-Source with 2 datasets} & Bu,Bo & 89.1 & \multirow{4}{*}{89.65} \\ \cline{2-3}
 & Bu,Li & 88.7 &  \\ \cline{2-3}
 & Bu,Os & 90.6 &  \\ \cline{2-3}
 & Bu,Wi & 90.2 &  \\ \hline
\multirow{6}{*}{Multi-Source with 3 datasets} & Bu,Bo,Wi & 91.3 & \multirow{6}{*}{91.95} \\ \cline{2-3}
 & Bu,Li,Os & 91.5 &  \\ \cline{2-3}
 & Bu,Li,Bo & 91.9 &  \\ \cline{2-3}
 & Bu,Li,Wi & 92.1 &  \\ \cline{2-3}
 & Bu,Os,Bo & 92.3 &  \\ \cline{2-3}
 & Bu,Os,Wi & 92.6 &  \\ \hline
\end{tabular}
\end{table}

\begin{table}[]
\centering
\caption{Comparison of classification accuracy and average accuracy (\%) on food packaging target dataset Bourne ($Bo$)}
\label{tab:my-table Bo}
\begin{tabular}{|l|l|l|l|}
\hline 
\textbf {Method} & \textbf {Source} & \textbf{Accuracy} & \textbf {Average} \\ \hline
Single-Source & \textbf{Ab} & 82.6 & 82.60 \\ \hline
\multirow{4}{*}{Source Combined  with 2 datasets} & Ab,Bu & 84.8 & \multirow{4}{*}{84.70} \\ \cline{2-3}
 & Ab,Li & 83.5 &  \\ \cline{2-3}
 & Ab,Os & 86.3 &  \\ \cline{2-3}
 & Ab,Wi & 84.2 &  \\ \hline
\multirow{6}{*}{Source Combined with 3 datasets} & Ab,Bu,Li & 86.2 & \multirow{6}{*}{85.68} \\ \cline{2-3}
 & Ab,Bu,Os & 84.9 &  \\ \cline{2-3}
 & Ab,Bu,Wi & 85.2 &  \\ \cline{2-3}
 & Ab,Li,Os & 86.9 &  \\ \cline{2-3}
 & Ab,Li,Wi & 84.1 &  \\ \cline{2-3}
 & Ab,Os,Wi & 86.8 &  \\ \hline
\multirow{4}{*}{Multi-Source with 2 datasets} & Ab,Bu & 90.2 & \multirow{4}{*}{90.70} \\ \cline{2-3}
 & Ab,Li & 91.1 &  \\ \cline{2-3}
 & Ab,Os & 90.3 &  \\ \cline{2-3}
 & Ab,Wi & 91.2 &  \\ \hline
\multirow{6}{*}{Multi-Source with 3 datasets} & Ab,Bu,Li & 94.2 & \multirow{6}{*}{92.75} \\ \cline{2-3}
 & Ab,Bu,Os & 92.6 &  \\ \cline{2-3}
 & Ab,Bu,Wi & 92.1 &  \\ \cline{2-3}
 & Ab,Li,Os & 92.3 &  \\ \cline{2-3}
 & Ab,Li,Wi & 92.9 &  \\ \cline{2-3}
 & Ab,Os,Wi & 92.4 &  \\ \hline
\end{tabular}
\end{table}

\begin{table}[]
\centering
\caption{Comparison of classification accuracy and average accuracy (\%) on food packaging target dataset Windmill-lane ($Wi$)}
\label{tab:my-table Wi}
\begin{tabular}{|l|l|l|l|}
\hline 
\textbf {Method} & \textbf {Source} & \textbf{Accuracy} & \textbf {Average} \\ \hline
Single-Source & \textbf{Ab} & 83.2 & 83.20 \\ \hline
\multirow{4}{*}{Source Combined with 2 datasets} & Ab,Bo & 84.5 & \multirow{4}{*}{83.27} \\ \cline{2-3}
 & Ab,Bu & 85.6 &  \\ \cline{2-3}
 & Ab,Li & 82.1 &  \\ \cline{2-3}
 & Ab,Os & 80.9 &  \\ \hline
\multirow{6}{*}{Source Combined with 3 datasets} & Ab,Bu,Li & 83.6 & \multirow{6}{*}{84.73} \\ \cline{2-3}
 & Ab,Bu,Os & 86.1 &  \\ \cline{2-3}
 & Ab,Bu,Bo & 86.3 &  \\ \cline{2-3}
 & Ab,Li,Os & 84.1 &  \\ \cline{2-3}
 & Ab,Li,Bo & 84.6 &  \\ \cline{2-3}
 & Ab,Os,Bo & 83.7 &  \\ \hline
\multirow{4}{*}{Multi-Source with 2 datasets} & Ab,Bo & 90.3 & \multirow{4}{*}{91.05} \\ \cline{2-3}
 & Ab,Bu & 89.5 &  \\ \cline{2-3}
 & Ab,Li & 91.6 &  \\ \cline{2-3}
 & Ab,Os & 92.8 &  \\ \hline
\multirow{6}{*}{Multi-Source with 3 datasets} & Ab,Bu,Li & 92.8 & \multirow{6}{*}{92.91} \\ \cline{2-3}
 & Ab,Bu,Os & 93.2 &  \\ \cline{2-3}
 & Ab,Bu,Bo & 92.7 &  \\ \cline{2-3}
 & Ab,Li,Os & 93.1 &  \\ \cline{2-3}
 & Ab,Li,Bo & 92.5 &  \\ \cline{2-3}
 & Ab,Os,Bo & 93.2 &  \\ \hline
\end{tabular}
\end{table}
As explained earlier, some of the benefits of applying the proposed approach to this application is the elimination of monotonous and inconsistent manual labor and reducing the human errors along with increasing speed and productivity. Firstly, we conducted experiments using labeled single source dataset and unlabeled single target dataset for all the six locations. The goal of this experiment has been to establish a baseline for images that would be classified as readable and acceptable according to human standards. Further experiments were conducted using our proposed multi-source domain adaptation approach this time, i.e. adapting two labeled source datasets with a single unlabeled target domain and three labeled source datasets with a single unlabeled target domain. We compared the obtained results with the baseline single source adaptation experiment conducted initially. Additional experiments were carried out by combining the two/three source datasets into a single source dataset. Our overall aim has been to improve the image classification accuracy in the provided food packaging datasets, hence allowing for enhanced quality control in the food supply chain. The combinations tested included all six locations available and were conducted in the following manner:

\begin{itemize}
     \item Single Source to Single Target
     \item Combined Source to Single Target
     \item Multi Source to Single Target 
\end{itemize}

We have trained our model on labeled data from a source domain to achieve better performance on data from a target domain, with access to only unlabeled data in the target domain. We used ResNet-50 \cite{he2016identity} pretrained on ImageNet \cite{russakovsky2015imagenet} as our backbone network and replaced the last fully connected (FC) layer with the task specific FC layer. We have fine-tuned all the convolutional and pooling layers and trained the classifier layer via back propagation. Adam optimizer with a learning rate of 0.001 was used \cite{kingma2014adam}.

\subsection{Single-Source Domain Adaptation}
The labeled source and the unlabeled target images have been fed through the model where the discrepancy between the pair of datasets was minimized by jointly reducing the feature discrepancy and the classification losses. We performed multiple experiments with each location as the target domain and presented the results  from table \ref{tab:my-table Os} through to table \ref{tab:my-table Wi}, listing the results of each approach from different sources.  

\subsection{Combined Source to Single Target}
In the source combined setting, all the source domains are combined into a single domain, and the experiments are conducted in a traditional single domain adaptation manner.
The labeled source and the unlabeled target images have been fed through the model where the discrepancy between the pair of datasets was minimized by jointly reducing the feature discrepancy and the classification loss. We have combined and experimented using
\begin{itemize}
     \item Two sources  combined 
     \item Three sources combined 
\end{itemize}

In the first set of experiments,  two datasets were combined into a single source dataset, whereas in the next set three source datasets were combined into a single source dataset. The results of this method can be seen from table \ref{tab:my-table Os} through to table \ref{tab:my-table Wi} with each location as the target dataset, listing the results of each approach from different sources. 

\subsection{Multi-Source Domain Adaptation}
The labeled sources and the unlabeled target images have been fed through the model where the discrepancy between the pair of datasets was minimized by jointly reducing the feature discrepancy, class discrepancy and the classification losses using the techniques described in section III. We performed multiple experiments per location as target domain and presented the results in tables \ref{tab:my-table Os} through to \ref{tab:my-table Wi} for each location. We categorized the experiments as follows:
\begin{itemize}
     \item Multi-Source with two datasets 
     \item Multi-Source with three datasets
\end{itemize}
Initially, we performed the experiments taking two source datasets as input domains and in the second set of experiments, we took three sources as input domains. The results of this method can be seen from table \ref{tab:my-table Os} through to table \ref{tab:my-table Wi} with each location as the target domain, listing the results of each classifier from different sources. 

From the experimental results we can make two clear observations: 
The results of the source combine method are comparatively better than the single source method which could have been the result of data enrichment, indicating that combining multiple source domains into single source domain is helpful in most of the tasks.
Our multi-source domain adaptation approach significantly outperforms two of the above mentioned baseline methods on most of the tasks with an average classification accuracy improvement by more than 6\%. We can also note that adding more sources and learning domain-invariant features for each source-target pairs  along with exploiting the domain-specific class boundary information significantly increased the average classification accuracy in dataset.

The overall average accuracy for the single-source approach is 84.14\%, two and three source combined approach is 85.05\% and 86.13\% respectively and finally our multi-source approach with two and three sources is 90.53 \% and 92.50\% respectively.

\begin{table}[]
\centering
\caption{Comparison of average classification accuracy for all methods}
\label{tab:my-table}
\begin{tabular}{|l|l|lll}
\cline{1-2}
\textbf{Method}                        & \textbf{Average} (\%) &  &  &  \\ \cline{1-2}
Single Source                 & 84.14   &  &  &  \\ \cline{1-2}
Two Sources Combined            & 85.05   &  &  &  \\ \cline{1-2}
Three Sources Combined         & 86.13   &  &  &  \\ \cline{1-2}
Multi-Source 2-source dataset  & 90.53   &  &  &  \\ \cline{1-2}
Multi-Source 3-source dataset & 92.50   &  &  &  \\ \cline{1-2}
\end{tabular}
\end{table}
\section{Conclusion and future Work}
In this paper we proposed a multi-source adaptation methodology that attempts to adapt and generalize information from one dataset to another by automating the verification of the use-by dates on food packaging datasets. 
The results shown in table \ref{tab:my-table} demonstrate that the accuracy of the food packaging classification improved significantly by adding each additional source and learning domain-invariant features along with exploiting the class alignment.
The proposed approach can also be applied on wider aspects of food package control, such as the verification of the allergen labeling, barcode, nutritional information and many more. 
Our future work will extend this study to a much larger dataset, consisting of about half a million food packaging images. In addition we aim at improving the domain adaptation approach via incorporating adversarial components. 

\section*{Acknowledgment}

We would like to thank OAL (Olympus Automation Limited) Company for providing us with the data used in this study.

\bibliographystyle{IEEEtran}
\bibliography{IJCNN}

\end{document}